\documentclass[10pt,twocolumn,letterpaper]{article}

\usepackage[pagenumbers]{cvpr} 

\definecolor{cvprblue}{rgb}{0.21,0.49,0.74}
\usepackage[pagebackref,breaklinks,colorlinks,allcolors=cvprblue]{hyperref}

\usepackage{algorithm}
\usepackage{ragged2e}
\usepackage{dsfont}

\usepackage{amssymb}
\usepackage{amsmath}
\usepackage{gensymb}
\usepackage{booktabs}
\usepackage{multirow}
\usepackage{bbding}
\usepackage{dcolumn}
\usepackage{ragged2e}
\usepackage{threeparttable}
\usepackage{tabularx}
\usepackage{algpseudocode}

\DeclareMathOperator*{\argmin}{arg\,min}

\def\paperID{***} 
\def\confName{CVPR}
\def\confYear{2026}

\title{See Through the Noise: Improving Domain Generalization \\ in Gaze Estimation}

\author{Yanming Peng, Shijing Wang, Yaping Huang\thanks{Corresponding author.}, Yi Tian\\
Beijing Key Laboratory of Traffic Data Mining and Embodied Intelligence, Beijing Jiaotong University\\
{\tt\small pengym@bjtu.edu.cn, shijingwang@bjtu.edu.cn, yphuang@bjtu.edu.cn, tianyi@bjtu.edu.cn}
}

\begin{document}
\maketitle
\begin{abstract}


Generalizable gaze estimation methods have garnered increasing attention due to their critical importance in real-world applications and have achieved significant progress. However, they often overlook the effect of label noise, arising from the inherent difficulty of acquiring precise gaze annotations, on model generalization performance. In this paper, we are the first to comprehensively investigate the negative effects of label noise on generalization in gaze estimation. Further, we propose a novel solution, called See-Through-Noise (SeeTN) framework, which improves generalization from a novel perspective of mitigating label noise. Specifically, we propose to construct a semantic embedding space via a prototype-based transformation to preserve a consistent topological structure between gaze features and continuous labels. 
We then measure feature-label affinity consistency to distinguish noisy from clean samples, and introduce a novel affinity regularization in the semantic manifold to transfer gaze-related information from clean to noisy samples. Our proposed SeeTN promotes semantic structure alignment and enforces domain-invariant gaze relationships, thereby enhancing robustness against label noise. Extensive experiments demonstrate that our SeeTN effectively mitigates the adverse impact of source-domain noise, leading to superior cross-domain generalization without compromising the source-domain accuracy, and highlight the importance of explicitly handling noise in generalized gaze estimation.
\end{abstract}    
\section{Introduction}
\label{sec:intro}



Gaze estimation techniques have been extensively applied in various fields, including human-computer interaction~\cite{kyto2018pinpointing}, virtual reality~\cite{burova2020utilizing, konrad2020gaze}, and medical analysis~\cite{castner2020deep}.
Recently, powered by deep learning technologies, appearance-based gaze estimation~\cite{schneider2014manifold,zhang2015appearance,biswas2021appearance} has achieved significant performance improvement. However, existing models often suffer from substantial performance degradation when deployed to unseen domains. To enhance generalization in real-world scenarios, recent works have focused on learning domain-invariant representations to handle environmental variations via adversarial learning~\cite{cheng2022puregaze,wang2019generalizing}, contrastive learning~\cite{wang2022contrastive}, consistency constraints~\cite{xia2025collaborative,bao2022generalizing}, \textit{etc.}

\begin{figure}[!t]
	\centering
    {\includegraphics[width=\linewidth]{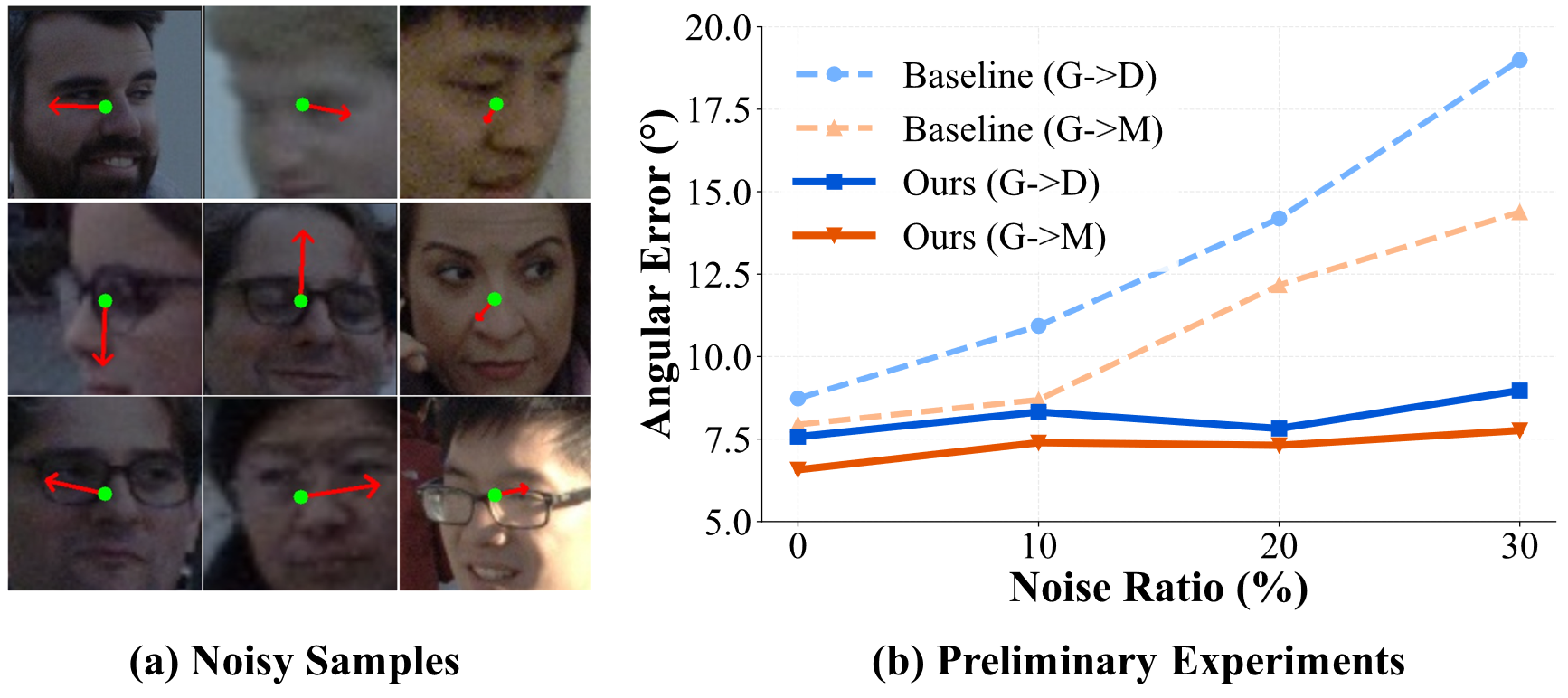}}
    
	\caption{(a) Noisy samples from the Gaze360 dataset, where red arrows indicate the provided ground-truth labels that clearly deviate from the true gaze directions. (b) Cross-domain results on the  synthetic noisy Gaze360 dataset under cross-domain settings, Eyediap (G→D) and MPIIGaze (G→M). The baseline (ResNet-18) performance degrades significantly with increasing noise, while our method consistently maintains robust generalization.} 
    \label{fig:noisy}
\end{figure}

Although these methods have achieved remarkable progress, they often overlook the widely existing noisy labels in the gaze estimation task. In practice, gaze estimation datasets are notoriously difficult to collect due to several factors, including unstable human attention and environmental variations (\textit{e.g.}, lighting, head pose, and occlusion). These challenges lead to inevitable annotation noise in large-scale gaze datasets.
As shown in Fig.~\ref{fig:noisy}(a), a considerable amount of noise exists in the commonly used Gaze360 dataset~\cite{kellnhofer2019gaze360}. 
Intuitively, overfitting to noisy labels forces the model to inadvertently learn gaze‑irrelevant features, thereby degrading its generalization performance. To comprehensively explore the impact of noisy labels in cross-domain scenarios, we conduct preliminary experiments on the synthetic noisy dataset by injecting label noise into the Gaze360 dataset. In detail, we randomly select 10$\%$, 20$\%$, and 30$\%$ of the samples and add Gaussian noise with a standard deviation of 60$\degree$ to their labels. As shown in Fig.~\ref{fig:noisy}(b), we can observe that label noise adversely affects the model’s generalization performance, \textit{e.g.}, the performance degrades from 7.94$\degree$ to 12.17$\degree$ under 20\% and further to 14.38$\degree$ under 30\% label noise on MPIIGaze. Moreover, this is also validated by the results of real-world experiments (see Sec.~\ref{Sec:result} for details).


Unfortunately, most existing noisy label learning methods mainly focus on mitigating within-domain performance degradation in classification tasks, but the challenge we aim to address is a regression task with cross-domain degradation. 
This raises two key issues that remain unexplored in related studies and thus need to be addressed. First, compared with classification tasks, dealing with label noise in regression is substantially more challenging due to the continuous nature of the target space. In regression, label noise is not categorical misclassification but continuous deviation, making it difficult to define clear boundaries between clean and noisy samples. Moreover, the feature space often lacks distinct clusters, and common classification-based heuristics such as small-loss selection or transition-matrix estimation become unreliable. Second, due to the inherent difficulty of cross-domain scenarios, simply correcting noisy labels does not necessarily improve generalization performance, because such correction primarily refines the label quality within the source domain but fails to bridge the distribution gap between the source and target domains.

In this paper, we address the domain generalization in gaze estimation from a novel perspective of mitigating noise. We propose a \textbf{See}-\textbf{T}hrough-\textbf{N}oise framework, called SeeTN, which models the semantic relatedness between continuous gaze labels and feature representations to effectively identify noisy samples and mitigate domain distribution discrepancies. This is particularly crucial for gaze regression tasks, where labels are inherently continuous. 
Specifically, we propose a manifold representation of the semantic space using a prototype-based transformation, which can preserve and constrain the relative relationships among semantic features according to the label affinity. Through the proposed semantic manifold, we can measure the discrepancy of features and labels, which greatly facilitates the distinction between noisy and clean samples. Built upon the separation, we further propose a novel affinity regularization in semantic manifold space, where different losses are respectively applied to clean and noisy samples, thereby resulting in more robust regularization of gaze-relevant features.
Comprehensive experimental results show that SeeTN effectively mitigates the negative impacts of noisy labels and improves the model's domain generalization ability. In conclusion, our contributions can be summarized in three aspects:

\begin{itemize}
    \item We present the first work to investigate and address the domain generalization of gaze estimation from a novel perspective of label noise, arising inevitably during the process of gaze dataset acquisition.
    
    \item We propose a novel See-Through-Noise framework, named SeeTN, which bridges continuous gaze angle labels and gaze features in a semantic manifold constructed via a prototype-based transformation. By leveraging the manifold space, we measure the discrepancy between feature and label affinities to identify noisy samples, and further design a novel affinity regularization strategy for robust learning of clean and noisy samples.
    

    \item We conduct comprehensive experiments on various cross-domain settings. The results on real-world datasets demonstrate that SeeTN effectively mitigates the negative impacts of noisy labels and learns more consistent semantic information.
    
\end{itemize}

\section{Related Works}
\label{sec:related_works}

\subsection{Cross-domain Gaze Estimation}

The challenge of enabling gaze estimation models trained in controlled laboratory settings to perform reliably in complex real-world scenarios has led to growing interest in cross-domain gaze estimation. Generally, cross-domain gaze estimation can be categorized into two settings: unsupervised domain adaptation and domain generalization.

Unsupervised domain adaptation assumes that only a limited amount of unlabeled target domain data is accessible. 
CRGA~\cite{wang2022contrastive} utilizes contrastive regression learning to acquire better feature representations. RUDA~\cite{bao2022generalizing} adapts to the target domain through rotation consistency of gaze. UnReGA~\cite{cai2023source} proposes source-free unsupervised domain adaptation in gaze estimation. 
Domain generalization, which cannot access target domain data, is more versatile and challenging. PureGaze~\cite{cheng2022puregaze} purifies gaze features to improve generalization performance. AGG~\cite{bao2024feature} mitigates the overfitting during the training stage by replacing the last FC layer. GazeCF~\cite{xu2023learning} leverages synthetic data to purify gaze-related features. CGaG~\cite{xia2025collaborative} employs collaborative contrastive learning to learn the gaze-related features. 

Although both adaptation and generalization methods have achieved remarkable performance, they have largely overlooked the potential noise in gaze labels of the source domain, which serve as the supervision signals and provide the domain-invariant semantic information in cross-domain gaze estimation. In contrast, we propose a novel perspective of handling noisy labels to enhance the domain generalization performance. 

\begin{figure*}[t]
    \centering
    \includegraphics[width=0.88\textwidth,keepaspectratio]{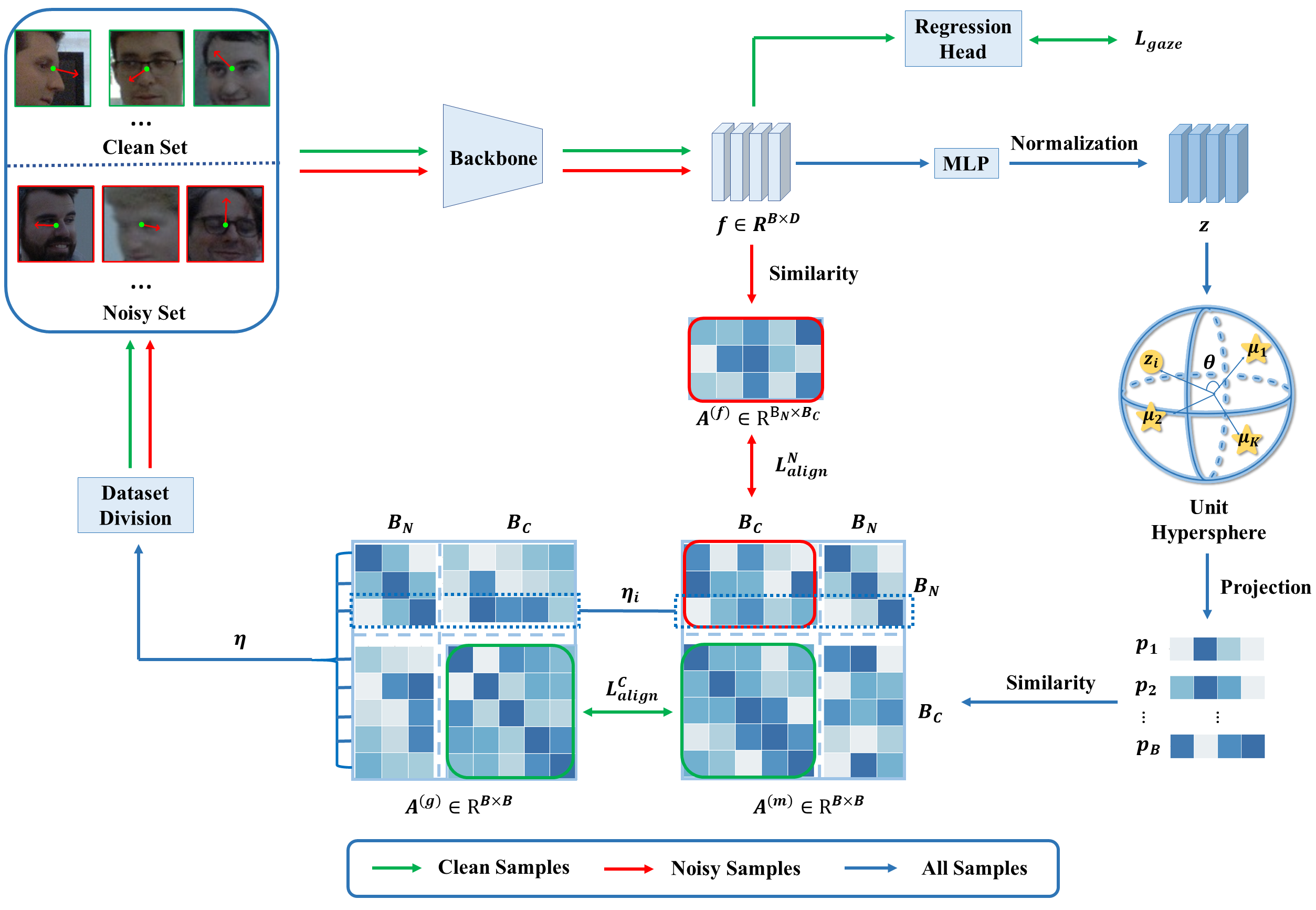}

    \caption{Overview of the proposed SeeTN, which consists of three main steps. First, SeeTN constructs the semantic manifold to strengthen the associations between gaze features and their corresponding continuous gaze labels. Second, according to the semantic discrepancies between features and their labels in the manifold, SeeTN divides the dataset into a noisy set and a clean set. Finally, to better utilize noisy samples and clean ones, SeeTN introduces a noise-robust regularization mechanism that constrains the noisy and clean sets separately. In the figure, red and green denote noisy and clean samples, respectively.}
    
    \label{fig:STN}
\end{figure*}

\subsection{Noisy Label Learning}

The mainstream methods of noisy label learning (LNL) that mainly focus on classification tasks can be roughly categorized into loss correction and sample selection. For loss correction, a common strategy is to estimate the noise transition matrix and design regularized loss functions~\cite{amid2019robust, xiao2015learning}. Recently, benefiting from the development of semi-supervised learning techniques, sample selection methods have become more popular. Many approaches~\cite{li2020dividemix,tu2023learning,wei2023fine} divide datasets into clean and noisy subsets via sample selection strategies and leverage semi-supervised learning to generate pseudo-labels and improve performance. Small-loss criterion~\cite{gui2021towards, zhang2023rankmatch, yu2019does} is commonly used in sample selection due to the memorization effect that DNNs are more likely to fit clean samples in early training. 

However, LNL is predominantly designed for classification tasks, whereas gaze estimation is a regression task. We have empirically demonstrated that simply transferring LNL techniques to gaze estimation is not enough. 
To the best of our knowledge, in gaze estimation, only SUGE~\cite{wang2024suppressing} first utilizes the co-training strategy, estimates the uncertainty for suppressing noisy samples, and generates pseudo-labels via neighbor labeling. Nevertheless, SUGE addresses only the within-domain setting, overlooking the more practical cross-domain scenario. In cross-domain settings, addressing both noise and generalization simultaneously for gaze estimation poses a significant challenge.

\section{Method}
\label{sec:method}

\subsection{Problem Definition}
We denote the source domain data as $\mathcal{D}_{\mathcal{S}} = \{(x_i, y_i)\}_{i=1}^{N_{\mathcal{S}}}$,
where $(x_i, y_i)$ corresponds to an image and its 3D gaze label. Let $F$ and $G$ denote the feature extractor and regressor of DNNs respectively, and the prediction of the image $x$ is $G(F(x;\theta_f);\theta_g)$, where $\theta_f$ and $\theta_g$ denote the parameters of the model. To simplify the formulations, we use $f(x)=F(x;\theta_f)$ and $g(x)=G(F(x;\theta_f);\theta_g)$ as alternatives. For domain generalization, typically, the objective function in the source domain can be formulated as:
\begin{equation}
    \begin{aligned}
        \theta^* = \argmin_{\theta} \mathbb{E}_{(x,y)} [\mathcal{L}_1(g(x), y)], 
    \end{aligned}
\label{equ:source}     
\end{equation}
and the model with $\theta^*$ is directly used to test the target domain. Since $x$ and $y$ contain low-quality images and labels, the model could overfit the noise, leading to poor domain generalization performance. 

\subsection{Overview}

Fig.~\ref{fig:STN} shows the overall framework of the proposed SeeTN. 
SeeTN aims to enhance the semantic association between gaze features and continuous gaze angles, which facilitates preserving the geometric structure of the semantic space in the gaze regression task. 
Meanwhile, the semantic discrepancies between labels and features observed during the association process can assist the model in identifying noisy samples.
To achieve the goal, SeeTN first constructs a semantic manifold from feature-derived prototypes, in which the affinity constraints between features in the manifold can be formulated based on the similarity between their corresponding labels. By quantifying the discrepancy of features and labels, an indicator is designed for identifying noisy labels. For noisy samples with unreliable labels, SeeTN regularizes the raw features by leveraging the affinity matrix in the manifold, which seeks to transfer the gaze-related information supervised by the clean samples to noisy samples. Through tailored regularization for clean and noisy samples, our model strengthens the learning of gaze-relevant representations, leading to the improvement of domain generalization. In the following sections, we present the proposed method in detail.

\subsection{Semantic Manifold Construction}

In regression tasks, the commonly used mean absolute error (MAE) loss focuses on the numerical accuracy of predictions, while overlooking the semantic structures in both the feature and label spaces. This implies that samples with similar labels may still have feature representations scattered across the feature space, which is detrimental to the semantic correspondence between features and labels. For better identifying noisy samples and improving model generalization, it is essential to strengthen the semantic alignment between features and labels. Thus motivated, we propose to explicitly construct a semantic manifold $\mathcal{M}$ to model the semantic structures of features.

Formally, to emphasize gaze semantics and suppress irrelevant information, we 
first transform the feature $f$ into another space through MLP and project it onto the unit hypersphere as ${z}=\textrm{Norm}(\textrm{MLP}(f))$. Then, $K$ prototypes, denoted as $\mu_k, k=1,2,\dots,K$, are randomly selected in the unit hypersphere to serve as the basis vectors of $\mathcal{M}$. To ensure that all prototypes $\mu_k$ contain more representative semantic information, they are progressively moved toward the feature center via an Exponential Moving Average (EMA) update:
\begin{gather}
    r = \textrm{Softmax}( z \cdot \mu^T/\tau), \\
    \mu_k \gets \alpha \ \mu_k + (1-\alpha)\frac{\sum_{i}r_{i,k}{z}_i}{\sum_{i}r_{i,k}},
\label{equ:mu_ema}
\end{gather}
where $r \in \mathbb{R}^{B \times K}$ denotes the assignment probabilities of $z$ with respect to the prototypes $\mu$, $B$ is the batch size, and $\alpha$ represents the EMA momentum and is set to 0.95. In addition, relying on the semantic prototypes $\mu$, the representation $p$ of $z$ on the manifold $\mathcal{M}$ can be expressed:
\begin{align}
& p  = {z} \cdot \mu^{\rm{T}}.
\label{equ:p}
\end{align}

Motivated by the continuous rather than categorical nature of gaze angles, we constrain the learned semantic manifold to preserve a topological structure consistent with that of the label space. Therefore, we first measure the affinity matrix in the manifold $\mathcal{M}$ denoted as $A^{(m)}$, which represents the relative relationships between samples defined by cosine similarity:
\begin{align}
& A^{(m)}_{i,j}  = \frac{p_{i} \cdot p_{j}}{\| p_{i} \| \|p_{j}\|}.
\label{equ:Am}
\end{align}
The gaze label affinity matrix $A^{(g)}$ can also be calculated by
\begin{align}
 A^{(g)}_{i,j}  = \frac{y_{i} \cdot y_{j}}{\| y_{i} \| \|y_{j}\|}.
 \label{equ:Ay}
\end{align}
Then we can align the gaze label affinity matrix $A^{(g)}$ and the manifold feature affinity matrix $A^{(m)}$ for establishing a bridge between them on the $\mathcal{M}$. By minimizing the MAE $|A^{(m)} - A^{(g)}|$ between manifold feature and label affinity matrix, 
the manifold $\mathcal{M}$ is encouraged to emphasize the semantic relationships between samples, while reinforcing the similarity-based associations between features and labels, thus providing a solid basis for the subsequent noise filtering and robust learning process. 

\subsection{Noise‑Robust Regularization}

In gaze estimation, noisy samples are inevitable, as collecting and annotating large-scale datasets is inherently challenging. To alleviate the negative impact of noisy samples and improve the generalization abilities, we propose to distinguish noisy samples from clean ones. 
Although cross entropy loss-based clean sample selection is not feasible for regression tasks, by exploiting the semantic manifold, we can quantify the affinity mismatch between features and labels using the cross-entropy function as a criterion. Moreover, we further design a novel affinity regularization to improve the model generalization.

Specifically, intuitively, the normalized row vector $\hat{y}_{i}^{(m)} = \textrm{Softmax}(A_{i,:}^{(m)})$ in the semantic manifold $\mathcal{M}$ can be interpreted as a distribution of sample $x_i$ based on its relations to the remaining samples. Similarly, the row vector $\hat{y}_i^{(g)}=\textrm{Softmax}(A_{i,:}^{(g)})$ can also be regarded as the distribution of sample $x_i$ over all other samples in the gaze label space. Therefore, for each sample $x_i$, we can introduce an indicator $\eta_i$ to describe the discrepancy between its feature and label as follows:
\begin{align}
& \eta_i = -\sum_{j=1}^{B} \hat{y}_{i,j}^{(g)} \log \hat{y}_{i,j}^{(m)}.
\label{equ:celoss}
\end{align}
By integrating features, labels and the relative relationships among samples, the indicator $\eta_i$ possesses the capability to identify noise. Noisy samples tend to exhibit inconsistencies between the pairwise relationships in the label space and those in the manifold feature space, therefore, samples with larger $\eta$ can be regarded as noisy samples.
To divide samples into clean and noisy ones, we adopt a simple strategy by sorting all samples in descending order according to $\eta$, and selecting the top $t\%$ as the noisy subset $\mathcal{D_S^N}$, while the remaining samples are regarded as the clean subset $\mathcal{D_S^C}$. 

Admittedly, this simple partitioning strategy may introduce risks. $\mathcal{D_S^N}$ might contain some hard samples that have not yet been well learned by the model, and $\mathcal{D_S^C}$ may still include a few noisy samples with small deviations. Nevertheless, this approach places most noisy samples and low‑quality images in the noisy subset, which is beneficial for improving model robustness. Meanwhile, before each training epoch, we repartition the dataset according to the updated $\eta$ to mitigate confirmation bias.

Based on the above partition, we further design a noise‑robust regularization item that applies distinct constraints to clean and noisy samples separately. For $\mathcal{D_S^C}$, as the labels are more reliable, we directly use them to supervise the gaze estimation loss and constrain the affinity matrix $A^{(m)}$  within the manifold space $\mathcal{M}$:
\begin{gather}
 \mathcal{L}_{\rm{gaze}}  = \frac{1}{B_{\scriptscriptstyle  C}}\sum_{x_{i} \in \mathcal{D_S^C} }| g(x_i) - y_{i} |,\\
 \mathcal{L}_{\rm{align}}^\mathcal{C}  = \frac{1}{B_{\scriptscriptstyle  C}(B_{\scriptscriptstyle  C}-1)} \sum_{x_{i}\in \mathcal{D_S^C}}\sum_{x_{j}\in\mathcal{D_S^C}, i\neq j} |A_{i,j}^{(g)} - A_{i,j}^{(m)}|,
\label{equ:L1loss}
\end{gather}
where $B_{\scriptscriptstyle  C}$ is the batch size of $\mathcal{D^C_S}$. 
For $\mathcal{D_S^N}$, a simple discard would lead to the loss of valuable information. Thus, we leverage the affinity matrix $A^{(m)}$, which is supervised by clean samples, to constrain the gaze feature affinity matrix $A^{(f)}$, thereby reducing dependence on low-quality labels and transferring clean supervisory signals to noisy samples:
\begin{gather}
    A_{i,j}^{(f)}  = \frac{f_{i}^{\mathcal{N}} \cdot f_{j}^{\mathcal{C}}}{\| f_{i}^{\mathcal{N}} \| \|f_{j}^{\mathcal{C}}\|}, \\
    \mathcal{L}_{\rm{align}}^{\mathcal{N}}  = -\frac{1}{B_{\scriptscriptstyle  N}}\sum_{x_{i}\in\mathcal{D_S^N}} \frac{A_{i,:}^{(f)} \cdot A_{i,:}^{(m)}}{\| A_{i,:}^{(f)}\| \|A_{i,:}^{(m)}\|},
\label{equ:x_sim}
\end{gather}
where $f_{i}^{\mathcal{N}}$ and $f_{j}^{\mathcal{C}}$ denote $f$ from $\mathcal{D_S^N}$ and $\mathcal{D_S^C}$ separately, $B_{\scriptscriptstyle N}$ is the batch size of $\mathcal{D_S^N}$. $A_{i,:}^{(f)}$ and $A_{i,:}^{(m)}$ denote the feature and manifold affinities between $x_i$ and clean samples in the batch respectively. Unlike Eq.~\ref{equ:L1loss}, which adopts a relatively rigid MAE loss to precisely align the affinity matrix $A^{(m)}$, Eq.~\ref{equ:x_sim} relaxes the constraint through row-wise cosine similarity. This soft alignment focuses on refining the pairwise relationships among features while maintaining their separability in the high-dimensional space.

From the perspective of noisy label learning, our proposed noise-robust regularization strategy makes full use of clean samples and performs feature-level semantic correction on noisy samples, thereby 
encouraging the model to learn domain-invariant features, which can enhance robustness to unseen domains.

\subsection{Overall}

In summary, our final objective loss function is:
\begin{align}
& \mathcal{L}_{\rm{all}}  = \mathcal{L}_{\rm{gaze}} + \mathcal{L}_{\rm{align}}^{\mathcal{C}}+\lambda\mathcal{L}_{\rm{align}}^{\mathcal{N}},
\label{equ:L_all}
\end{align}
where $\lambda$ is set to 0.1.
During inference in unseen domains, SeeTN utilizes solely the backbone to extract features, which are then fed into the regression head to predict the final gaze angles.



\section{Experiments}
\label{sec:experiments}

\begin{table*}[t]
\centering
\caption{Performance comparison of SeeTN with different backbones (ResNet-18 and ResNet-50) under four cross-domain gaze estimation settings and within-domain settings. Results are shown as mean angular errors ($\degree$), with red values denoting relative improvement over the baseline. Lower values indicate better performance. }
\begin{tabularx}{\textwidth}{l|>{\centering\arraybackslash}X>{\centering\arraybackslash}X|>{\centering\arraybackslash}X>{\centering\arraybackslash}X|>{\centering\arraybackslash}X>{\centering\arraybackslash}X}
\toprule[1pt]
Method &
$\mathcal{D}_E \!\to\! \mathcal{D}_M$ &
$\mathcal{D}_E \!\to\! \mathcal{D}_D$ &
$\mathcal{D}_G \!\to\! \mathcal{D}_M$ &
$\mathcal{D}_G \!\to\! \mathcal{D}_D$ &
within $\mathcal{D}_E^{*}$ &
within $\mathcal{D}_G^{*}$ \\
\midrule
ResNet-18      & 8.07 & 8.78 & 7.94 & 8.73 & 4.64 & 11.14 \\
ResNet-18+SeeTN  & 
\textbf{6.58} {\color{red} $\downarrow$18.5\%} &
\textbf{7.18} {\color{red} $\downarrow$18.2\%}  &
\textbf{6.57} {\color{red} $\downarrow$17.2\%}  &
\textbf{7.57} {\color{red} $\downarrow$13.3\%} &
\textbf{4.40} {\color{red} $\downarrow$5.1\%} &
\textbf{10.73} {\color{red} $\downarrow$3.7\%} \\
\midrule
ResNet-50      & 7.64 & 8.39 & 7.68 & 8.65 & 4.32 & 10.78 \\
ResNet-50+SeeTN  &
\textbf{6.31} {\color{red} $\downarrow$17.4\%}  &
\textbf{6.84} {\color{red} $\downarrow$18.4\%}  &
\textbf{6.75} {\color{red} $\downarrow$12.1\%}  &
\textbf{7.42} {\color{red} $\downarrow$14.2\%} &
\textbf{4.16} {\color{red} $\downarrow$3.7\%} &
\textbf{10.45} {\color{red} $\downarrow$3.1\%} \\
\bottomrule[1pt]
\end{tabularx}
\label{tab:compare_backbone}
\end{table*}

\begin{table}[t]
\centering
\small
\caption{Comparison of domain generalization performance between SeeTN and state-of-the-art gaze estimation methods. $*$ indicates methods with ResNet-50 backbone.}
\setlength{\tabcolsep}{1pt}
\begin{tabular}{l|cc|cc}
\toprule[1pt]
Method &
$\mathcal{D}_E \!\to\! \mathcal{D}_M$ &
$\mathcal{D}_E \!\to\! \mathcal{D}_D$ &
$\mathcal{D}_G \!\to\! \mathcal{D}_M$ &
$\mathcal{D}_G \!\to\! \mathcal{D}_D$ \\
\midrule
Full-Face~\cite{zhang2017s} & 12.35 & 30.15 & 11.13 & 14.42 \\
ADL~\cite{wang2019generalizing}       & 7.23  & 8.02  & 11.36 & 11.86 \\
CA-Net~\cite{cheng2020coarse}     & --    & --    & 27.13 & 31.41 \\
LatentGaze~\cite{lee2022latentgaze} & 7.98  & 9.81  & --    & --    \\
PureGaze~\cite{cheng2022puregaze}   & 7.08$^*$ & 7.48$^*$ & 9.28 & 9.32 \\
GazeCF~\cite{xu2023learning}  & 6.50 & 7.44 & 7.55 & 9.00 \\  
CGaG~\cite{xia2025collaborative}    & 6.47  & 7.03  & 7.50 & 8.67 \\
AGG~\cite{bao2024feature}    & 7.10  & 7.07  & 7.87 & 7.93 \\
AGG$^*$~\cite{bao2024feature} & \underline{5.91} & \textbf{6.75} & 9.20 & 11.36 \\
FSCI~\cite{liang2024confounded} & \textbf{5.79} & 6.96 & 7.06 & 7.99 \\
\midrule
ResNet18+SeeTN  & 6.58 & 7.18 & \textbf{6.57} & \underline{7.57}  \\
ResNet50+SeeTN$^*$  & 6.31 & \underline{6.84} & \underline{6.75} & \textbf{7.42}  \\
\bottomrule[1pt]
\end{tabular}
\label{tab:compare_generalization}
\end{table}


\subsection{Dataset Preparation}
We conduct experiments on four widely used gaze estimation datasets: ETH-XGaze ($\mathcal{D}_E$)~\cite{zhang2020eth}, Gaze360 ($\mathcal{D}_G$)~\cite{kellnhofer2019gaze360}, MPIIFaceGaze ($\mathcal{D}_M$)~\cite{zhang2017mpiigaze}, and EyeDiap ($\mathcal{D}_D$)~\cite{funes2014eyediap}. \textbf{ETH-XGaze}: The images are captured by high‑resolution cameras in a laboratory environment, covering a wide gaze range. \textbf{Gaze360}: The images are captured by a 360$\degree$ camera in streets. We use 84,902 images with frontal faces in our experiments. \textbf{MPIIFaceGaze}: The images are captured by web camera of laptop computers. \textbf{EyeDiap}: The images are captured under laboratory environment with screen and floating targets. For a fair comparison, we preprocess data following the same procedures as previous studies~\cite{cheng2024appearance}, and report performance  by using $\mathcal{D}_E$ and $\mathcal{D}_G$ as the source domains separately, with the remaining datasets serving as the target domains.


\subsection{Implementation Details}
SeeTN is implemented in PyTorch with ResNet-18~\cite{he2016identity} and ResNet-50~\cite{he2016identity} as the backbone. The model is trained using the Adam optimizer with a learning rate of 0.0001 for 100 epochs on Gaze360 and for 10 epochs on ETH-XGaze. During the early stage of training, we warm up the model for 10 epochs on Gaze360 and for 2 epochs on ETH-XGaze to obtain a reliable indicator $\eta$ for partitioning the dataset. Besides, we set the number of prototypes to $K=12$. In all the settings, batch sizes are set to 128. 

\subsection{Quantitative Evaluation}
\label{Sec:result}
\subsubsection{Effectiveness of SeeTN}

We first evaluate SeeTN with different backbones to verify its effectiveness in mitigating the effects of noisy labels and enhancing the model’s domain generalization capability. As shown in Tab.~\ref{tab:compare_backbone}, we conduct experiments in four cross-domain and two within-domain settings. Specifically, when integrated with ResNet-18 and ResNet-50, SeeTN consistently improves the performance across all scenarios. SeeTN achieves improvements of 18.5\% on $\mathcal{D}_E \!\to\! \mathcal{D}_M$ and 17.2\% on $\mathcal{D}_G \!\to\! \mathcal{D}_M$, without using any target-domain data. Moreover, many generalization approaches~\cite{bao2024feature} improve performance on unseen domains at the cost of degraded within-domain accuracy. In contrast, even under the within-domain settings $\mathcal{D}_E^*$ and $\mathcal{D}_G^*$, SeeTN still brings a 3–5\% reduction in error. These results sufficiently demonstrate that SeeTN effectively alleviates the negative influence of noisy labels while substantially improving the model’s robustness and generalization ability across different domains.

\begin{table}[t]
\centering
\small
\caption{Comparison of domain generalization performance between SeeTN and representative noisy-label learning methods.}
\setlength{\tabcolsep}{3pt}
\begin{tabular}{l|cc|cc}
\toprule[1pt]
Method &
$\mathcal{D}_E \!\to\! \mathcal{D}_M$ &
$\mathcal{D}_E \!\to\! \mathcal{D}_D$ &
$\mathcal{D}_G \!\to\! \mathcal{D}_M$ &
$\mathcal{D}_G \!\to\! \mathcal{D}_D$ \\
\midrule
Baseline & 8.07 & 8.78 & 7.94 & 8.73 \\
DivideMix~\cite{li2020dividemix} & 9.72 & 20.09 & 10.46 & 12.30 \\
SUGE~\cite{wang2024suppressing}      & 10.00  & 8.74  & 7.04 & 8.32 \\
\midrule
SeeTN    & \textbf{6.58} & \textbf{7.18} & \textbf{6.57} & \textbf{7.57} \\
\bottomrule[1pt]
\end{tabular}
\label{tab:compare_noisy}
\end{table}

\subsubsection{Comparison with Domain Generalization Methods}

We compare the domain generalization ability of the proposed SeeTN framework with several recent gaze estimation methods~\cite{zhang2017s,wang2019generalizing,cheng2020coarse,lee2022latentgaze,cheng2022puregaze,xu2023learning,xia2025collaborative,bao2024feature, liang2024confounded}. As shown in Tab.~\ref{tab:compare_generalization}, when trained on $\mathcal{D}_G$ as the source domain, SeeTN significantly outperforms other state-of-the-art methods, surpassing AGG~\cite{bao2024feature} by 2.63 and FSCI~\cite{liang2024confounded} by 0.49. In contrast, when trained on $\mathcal{D}_E$ as the source domain, its performance is slightly inferior to that of AGG$^*$ and FSCI. One possible reason, as also discussed in AGG~\cite{bao2024feature}, is the instability of the baseline model under cross-domain settings. Another reason is that $\mathcal{D}_G$ is more noisy than $\mathcal{D}_E$, which further indicates that the performance gains of SeeTN mainly stem from its explicit handling of noisy labels. More analysis of SeeTN’s ability to cope with noisy labels is provided in the supplementary material.



\subsubsection{Comparison with Noisy Label Learning Methods}

Since SeeTN improves domain generalization from the perspective of noisy labels, we compare it with DivideMix~\cite{li2020dividemix} and SUGE~\cite{wang2024suppressing} under cross-domain settings. DivideMix is a representative method in LNL, and extensive empirical evidence has verified its effectiveness in handling label noise. However, as shown in Tab.~\ref{tab:compare_noisy}, DivideMix performs poorly in gaze estimation and even underperforms the backbone, since it leverages MixMatch, a semi-supervised learning strategy designed for classification tasks, which is not well suited for regression problems. In gaze estimation, SUGE focuses on suppressing noisy labels in within-domain settings. We implement SUGE for cross-domain settings. It can be observed that SUGE achieves a slight improvement on $\mathcal{D}_G$, but its performance on $\mathcal{D}_E$ is unsatisfactory. SeeTN achieves significantly better performance than two previous methods, demonstrating the adaptability of our approach compared with existing LNL methods for domain generalization in gaze estimation.


\begin{table}[t]
\centering
\small
\caption{Ablation study for the effectiveness of each loss function on cross-domain settings.}
\setlength{\tabcolsep}{1pt}
\begin{tabular}{ccc|cc|cc}
\toprule[1pt]
$L_{\rm{gaze}}$ & $L_{\rm{align}}^{\mathcal{C}}$ & $L_{\rm{align}}^{\mathcal{N}}$ &
$\mathcal{D}_E \!\to\! \mathcal{D}_M$ &
$\mathcal{D}_E \!\to\! \mathcal{D}_D$ & 
$\mathcal{D}_G \!\to\! \mathcal{D}_M$ &
$\mathcal{D}_G \!\to\! \mathcal{D}_D$  \\
\midrule
\Checkmark &  &  & 8.07 & 8.78 & 7.94 & 8.73 \\
\Checkmark & \Checkmark &   & 7.14 & 7.95 & 7.26 & 8.11 \\
\Checkmark & \Checkmark & \Checkmark & \textbf{6.58} & \textbf{7.18} & \textbf{6.57} & \textbf{7.57} \\
\bottomrule[1pt]
\end{tabular}
\label{tab:ablation_loss_strategy}
\end{table}

\begin{table}[t]
\centering
\small
\caption{The ablation of different parameters and indicators on cross-domain scenarios.}
\setlength{\tabcolsep}{1.5pt}
\begin{tabular}{cc|cc|cc}
\toprule[1pt]
Params. & Value & $\mathcal{D}_E \!\to\! \mathcal{D}_M$ &
$\mathcal{D}_E \!\to\! \mathcal{D}_D$ &
$\mathcal{D}_G \!\to\! \mathcal{D}_M$ &
$\mathcal{D}_G \!\to\! \mathcal{D}_D$ \\
\midrule
 & 8 & 7.37 &  7.32 & 6.90 & \textbf{7.36} \\
$K$    & 12 & \textbf{6.58} & \textbf{7.18} & 6.57 & 7.57\\
    & 16 & 7.59 & 7.83 & \textbf{6.44} & 8.25 \\
\midrule
 & 5 & \textbf{6.58} & \textbf{7.18} & 6.85 & 7.93 \\
$t\%$ & 10 & 7.25 & 7.42 & \textbf{6.57} & \textbf{7.57} \\
  & 20 & 7.63 & 7.95 & 6.97 & 8.26\\
\midrule
\multirow{2}{*}{Indicator}   & L1 & 7.73 & 7.64 & 7.25 &  7.91 \\
 & $\eta$ & \textbf{6.58} & \textbf{7.18} & \textbf{6.57} & \textbf{7.57} \\
\bottomrule[1pt]
\end{tabular}
\label{tab:para}
\end{table}

\subsection{Ablation Study}

\noindent\textbf{Effect of Loss Functions.} We conduct ablation experiments to verify the effectiveness of each loss function designed in SeeTN. As shown in Tab.~\ref{tab:ablation_loss_strategy}, removing
$\mathcal{L}_{\rm{align}}^{\mathcal{N}}$ leads to a significant performance reduction of 0.56$\degree$ ($\mathcal{D}_E \!\to\! \mathcal{D}_M$), 0.77$\degree$ ($\mathcal{D}_E \!\to\! \mathcal{D}_D$), 0.69$\degree$ ($\mathcal{D}_G \!\to\! \mathcal{D}_M$) and 0.54$\degree$ ($\mathcal{D}_G \!\to\! \mathcal{D}_D$).
Removing $\mathcal{L}_{\rm{align}}^{\mathcal{C}}$ results in a further performance decline.
It can be concluded that both $\mathcal{L}_{\rm{align}}^{\mathcal{C}}$ and $\mathcal{L}_{\rm{align}}^{\mathcal{N}}$ improve the performance on all the cross-domain settings compared with the baseline, showing the significance of mitigating and utilizing noisy labels in domain generalization for gaze estimation.

\noindent\textbf{Number of Prototypes $\mu$ ($K$).} To explore the effect of the number of $\mu$ on SeeTN, we conduct experiments with different $\mu$ values under the cross-domain settings. As shown in Tab.~\ref{tab:para}, the performance shows a decreasing trend as $K$ increases. We infer that, due to the lack of explicit semantic distinction among prototypes in regression tasks, a larger $K$ makes the prototypes $\mu$ more likely to be affected by within-domain styles, leading to a decline in performance. Overall, $K=12$ is an appropriate choice for the source domain.

\noindent\textbf{Ratio of Divided Noisy Samples ($t\%$).} In SeeTN, we regard the top $t\%$ of samples with larger $\eta$ values as noisy samples. To explore the sensitivity of our method to $t$, we conduct experiments with various values of $t$, as shown in Tab.~\ref{tab:para}. In $\mathcal{D}_E$, we select $t\%=5\%$ since the dataset is collected in a laboratory environment with lower noise ratio. In $\mathcal{D}_G$, we select $t\%=10\%$ for Gaze360. As $t$ increases, the performance shows a slight decrease, but it still remains considerably higher than that of the baseline.

\noindent\textbf{Different Indicators.} For gaze estimation tasks, we specially design a novel indicator $\eta$ to distinguish noisy samples by pairwise relations among samples. To verify the effectiveness of $\eta$, we compare different indicators on cross-domain settings. Specifically, based on the small-loss criterion in noisy label learning, we replace $\eta$ with L1 loss to identify noisy samples. As shown in Tab.~\ref{tab:para}, SeeTN with $\eta$ achieves better performance than the L1 loss, showing the reliability of the proposed noisy detector. More ablations are provided in the supplementary material.

\subsection{Qualitative Evaluation}

\begin{figure*}[t]
    \centering
    \subfloat[Feature Visualization of Baseline]{
    \includegraphics[width=0.41\linewidth,keepaspectratio]{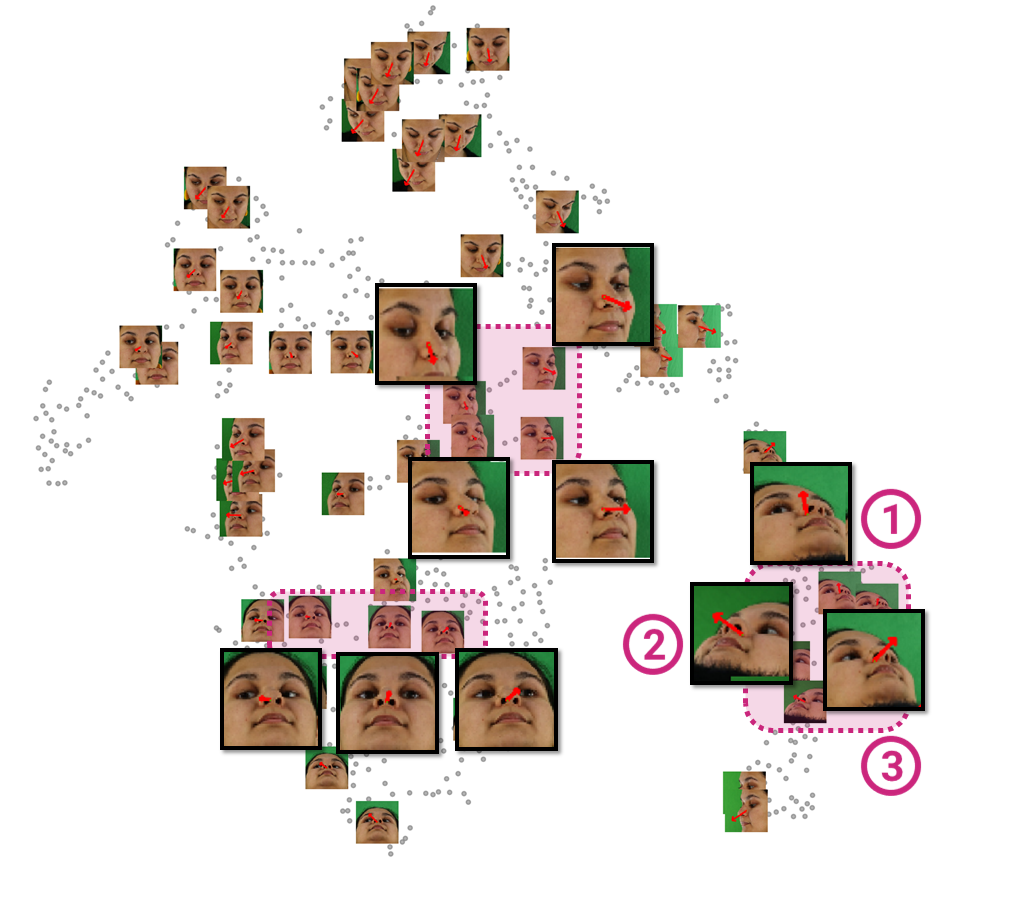}}
    \subfloat[Feature Visualization of SeeTN]{
    \includegraphics[width=0.51\linewidth,keepaspectratio]{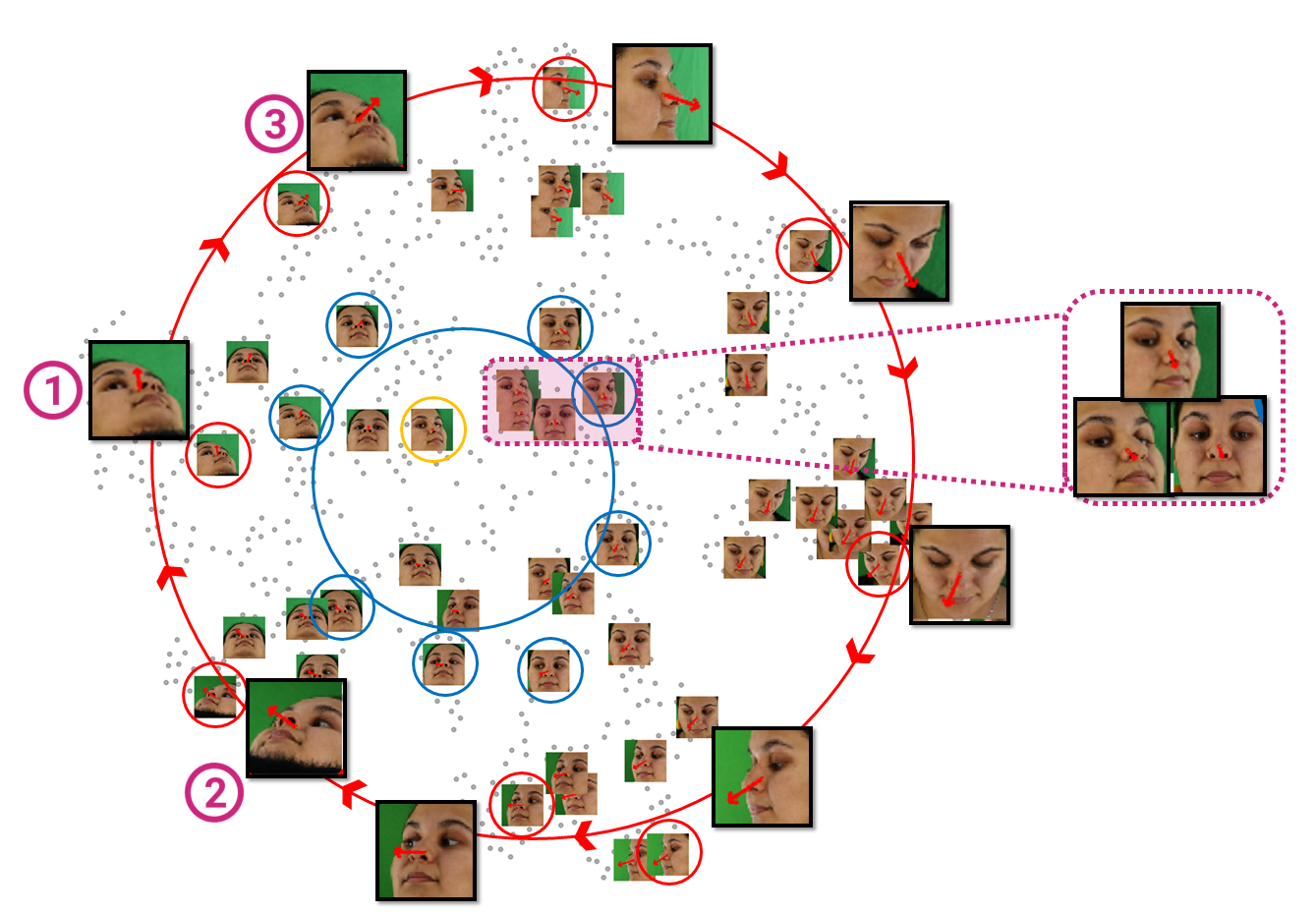}}
    \caption{The qualitative results of visualizing the features learned by the backbone and SeeTN on the ETH-XGaze dataset via t-SNE. The red arrows in the images denote the label directions of the samples.}
    \label{fig:tsne}
\end{figure*}

\subsubsection{Semantic Feature Visualization}

To qualitatively evaluate the effect of SeeTN on the feature space, we randomly select 800 images from one subject and visualize their feature representations under the baseline and our SeeTN using t‑SNE~\cite{maaten2008visualizing}.

As shown in the pink dashed regions of Fig.~\ref{fig:tsne}(a), it is evident that the baseline clusters the features mainly according to head pose rather than the more task-relevant gaze direction, which indicates that the model has not truly learned gaze‑related semantics, and thus its cross-domain performance obviously decreases when the domain style changes.

Encouragingly, as shown in Fig.~\ref{fig:tsne}(b), the features learned by our SeeTN are strongly associated with the gaze direction rather than head pose, as illustrated by the pink dashed regions. From the visualization results, we observe that the feature distribution learned by SeeTN exhibits the following three properties: 
\begin{itemize}
    \item The clustering of sample features is mainly based on the gaze direction rather than the head pose. For instance, in the baseline visualization, three samples clustered in the lower‑right corner share similar head poses, whereas in SeeTN, they are dispersed to different positions in the feature space according to their respective gaze directions. 
    \item The labels corresponding to the sample features change continuously in the feature space, which can be illustrated by the samples located on the red and blue circles in Fig.~\ref{fig:tsne}(b). It can be observed that, in the t‑SNE visualization of SeeTN, the gaze directions corresponding to these samples change continuously in a clockwise manner, demonstrating that our approach effectively preserves the inherent continuity of labels in gaze regression tasks.
    \item The deviation of the sample gaze direction from the frontal direction increases radially outward. For example, the gaze directions at corresponding radial positions on the blue and red circles exhibit nearly identical in the images, whereas the label arrows become progressively longer toward the periphery. The longer the arrow is, the larger the deviation angle is between the gaze direction and the frontal direction. This further indicates that the features learned by our SeeTN exhibit semantic continuity with respect to the gaze labels.   
\end{itemize}

The aforementioned observations and analyses clearly validate the consistency between the motivation, design methodology, and experimental results of the proposed SeeTN. By enforcing that similar features exhibit similar gaze directions within a semantic manifold space, we embed semantic information into the feature representation and enhance the association between features and gaze direction. This design reduces the model’s reliance on domain-specific style information, thereby improving its generalization ability across domains.

\subsubsection{Visualization of Noisy Sample Detection}

\begin{figure}[t]
    \centering
    \includegraphics[width=0.9\linewidth,keepaspectratio]{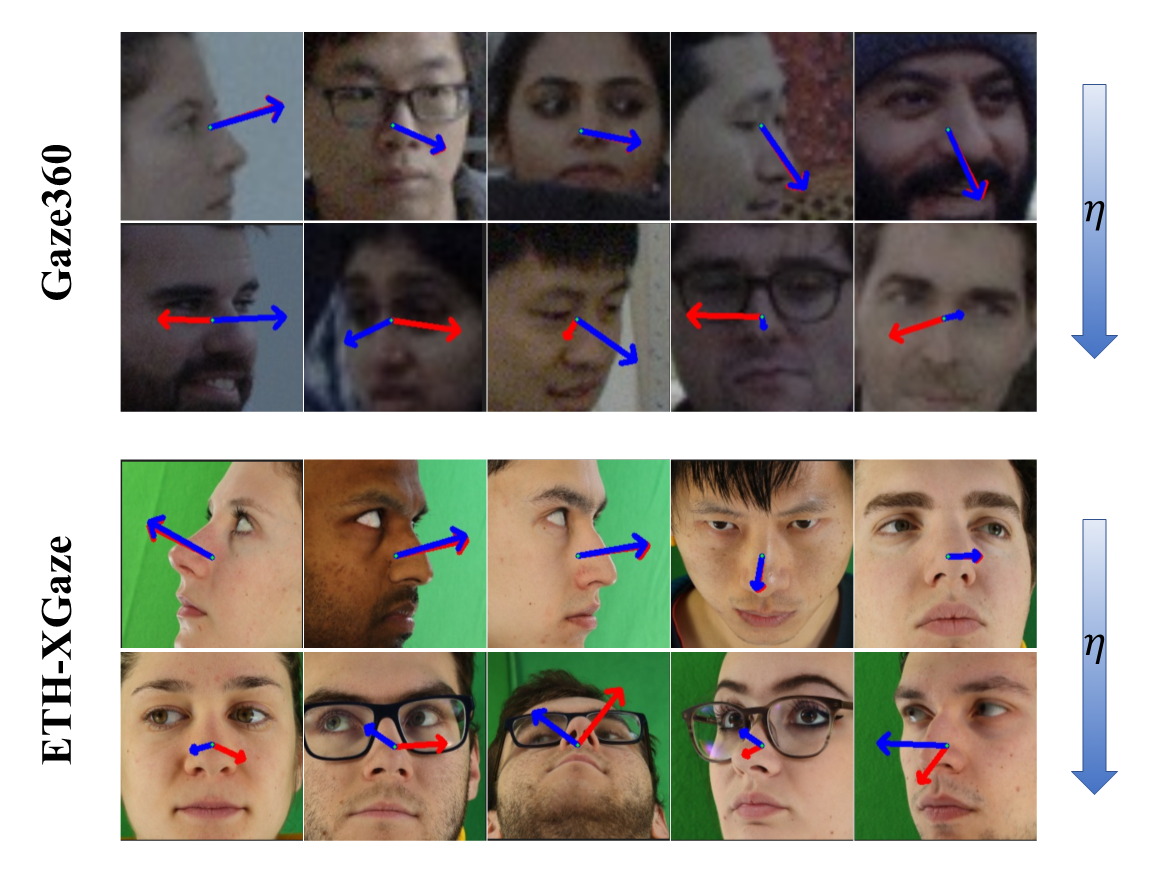}

    \caption{Sample Visualization  with the smallest and largest $\eta$ values in Gaze360 and ETH-XGaze. The red and blue arrows indicate the ground-truth directions and the model predictions respectively.}
    \label{fig:visual_samples}
\end{figure}

To intuitively illustrate how the indicator $\eta$ evaluates the samples, we visualize a subset of samples with the smallest $\eta$ values and those with the largest $\eta$ values. As shown in Fig.~\ref{fig:visual_samples}, for samples with low $\eta$ values, the labels are accurate, and the model’s predicted directions nearly coincide with the gaze labels. For samples with large $\eta$ values, their labels are often incorrect, while the model can still accurately predict their gaze directions. These results clearly demonstrate that the indicator $\eta$ can effectively assess whether a sample contains noisy labels, and also show that our model is robust to label noise. Additional visualization results can be found in the supplementary material.

\section{Conclusion and Discussion}
\label{sec:conclusion}

In this paper, we propose an effective framework named SeeTN to improve domain generalization performance in gaze estimation from a novel perspective of mitigating the negative impact of noisy labels. We first design a semantic manifold to preserve relationships between features and labels. We then measure feature-label affinity consistency to identify noisy samples, and introduce a noise-robust regularization in the semantic manifold to transfer gaze-related information from clean to noisy samples. Compared with existing methods, SeeTN achieves superior domain generalization performance without sacrificing within-domain accuracy, demonstrating the effectiveness of handling noisy samples in cross-domain gaze estimation task.

\textbf{Limitation.} While SeeTN has achieved significant improvement for various cross-domain settings, one limitation is that the strategy used to distinguish between clean and noisy samples is relatively simple. We will investigate more effective approaches that can accurately divide the noisy subset by combining noisy label learning methods with the semantic manifold proposed in this paper.

\section*{Acknowledgements}
\label{sec:Acknowledgements}

This work is supported by National Natural Science Foundation of China (62271042, 62376021) and Beijing Natural Science Foundation (L259055).

{
    \small
    \bibliographystyle{ieeenat_fullname}
    \bibliography{main}
}

\clearpage
\setcounter{page}{1}
\setcounter{section}{0}
\setcounter{table}{0}
\setcounter{figure}{0}
\maketitlesupplementary

\begin{table*}[t]
\centering
\normalsize
\caption{Comparison of domain generalization performance between SeeTN and representative noisy label learning methods in synthetic noisy Gaze360 dataset.}
\setlength{\tabcolsep}{5pt}
\begin{tabular}{l|cc|cc|cc|cc}
\toprule[1pt]
\multirow{2}{*}{Method} & \multicolumn{2}{c|}{Noise Ratio=0\%}& \multicolumn{2}{c|}{Noise Ratio=10\%}& \multicolumn{2}{c|}{Noise Ratio=20\%}& \multicolumn{2}{c}{Noise Ratio=30\%}\\
 & $\mathcal{D}_G \!\to\! \mathcal{D}_M$ &
$\mathcal{D}_G \!\to\! \mathcal{D}_D$ & $\mathcal{D}_G \!\to\! \mathcal{D}_M$ & $\mathcal{D}_G \!\to\! \mathcal{D}_D$ & $\mathcal{D}_G \!\to\! \mathcal{D}_M$& $\mathcal{D}_G \!\to\! \mathcal{D}_D$& $\mathcal{D}_G \!\to\! \mathcal{D}_M$& $\mathcal{D}_G \!\to\! \mathcal{D}_D$\\
\midrule
Baseline & 7.94 & 8.73 & 8.69 & 10.93 & 12.17 & 14.19 & 14.38 & 18.99\\
DivideMix & 10.46 & 12.3 & 10.32& 11.27 & 10.05 & 10.13 & 9.83 & 11.32\\
SUGE     & 7.04 & 8.32 & 7.43 & 8.61 & 7.40 & 9.12 & 9.23 & 11.37\\
\midrule
SeeTN  & \textbf{6.57} & \textbf{7.57}& \textbf{7.39} & \textbf{8.32} & \textbf{7.31} & \textbf{7.82} & \textbf{7.76} & \textbf{8.97}\\
\bottomrule[1pt]
\end{tabular}
\label{tab:synthetic_noise}
\end{table*}

\begin{table*}[t]
\centering
\normalsize
\caption{Comparison of domain generalization performance between SeeTN and AGG in a validation set with synthetic noise.}
\begin{tabular}{l|cc|cc|cc|cc}
\toprule[1pt]
\multirow{2}{*}{Method} & \multicolumn{2}{c|}{Noise Ratio=0\%}& \multicolumn{2}{c|}{Noise Ratio=10\%}& \multicolumn{2}{c|}{Noise Ratio=20\%}& \multicolumn{2}{c}{Noise Ratio=30\%}\\
 & $\mathcal{D}_E \!\to\! \mathcal{D}_M$ &
$\mathcal{D}_E \!\to\! \mathcal{D}_D$ & $\mathcal{D}_E \!\to\! \mathcal{D}_M$ & $\mathcal{D}_E \!\to\! \mathcal{D}_D$ & $\mathcal{D}_E \!\to\! \mathcal{D}_M$& $\mathcal{D}_E \!\to\! \mathcal{D}_D$& $\mathcal{D}_E \!\to\! \mathcal{D}_M$& $\mathcal{D}_E \!\to\! \mathcal{D}_D$\\
\midrule
Baseline & 10.25 & 11.04 & 13.02 & 13.53 & 14.23 & 15.01 & 15.33 & 16.72 \\
AGG & 9.59 & \textbf{8.84} & 10.52 & 10.41 & 12.19 & 13.21 & 13.96 & 18.06 \\
\midrule
SeeTN     & \textbf{8.57} & 9.37 & \textbf{9.10} & \textbf{9.82} & \textbf{8.65} & \textbf{9.68} & \textbf{8.91} & \textbf{10.05} \\
\bottomrule[1pt]
\end{tabular}
\label{tab:synthetic_noise2}
\end{table*}

In the supplementary material, we provide detailed description of our proposed SeeTN and more experiments results. We first present the training algorithm in Algorithm~\ref{alg:SeeTN}. Then we show additional quantitative experiments, including detailed results on synthetic noisy dataset, fine-tuned results on target domain and more hyperparameter analysis. Finally, we give more visualized results, including feature distribution of unseen domain and noisy sample detection. These results clearly verify the effectiveness of our proposed SeeTN approach.

\section{SeeTN Algorithm}

The full algorithm for implementing SeeTN is shown in Algorithm~\ref{alg:SeeTN}.

\begin{algorithm}
\caption{SeeTN}
\label{alg:SeeTN}
\begin{algorithmic}[1]
    \State {\textbf{Input:} Network $F(\cdot,\theta_f)$, Regressor $G(\cdot,\theta_g)$, Dataset $\mathcal{D_S}$, Prototypes $\mu$.}
    \State {$\theta_f, \theta_g = \rm{WarmUp}(G(F(\cdot,\theta_f),\theta_g))$}
    \State {Get initial $\mathcal{D_S^C}$ and $\mathcal{D_S^N}$ by $\eta$}
    
    \While {$t < \rm{MaxEpoch}$}
        \For {$\rm{item}=1$ \textbf{to} $\rm{num\_iters}$}
            \State  {From $\mathcal{D_S^C}$, draw a mini-batch$\{(x^\mathcal{C}_{i},y^\mathcal{C}_{i}), i=1...B_{\scriptscriptstyle  C}\}$}
            \State  {From $\mathcal{D_S^N}$, draw a mini-batch$\{(x^\mathcal{N}_{i},y^\mathcal{N}_{i}), i=1...B_{\scriptscriptstyle  N}\}$}
            \State {Get $f_i^\mathcal{C},f_i^\mathcal{N},g_i^\mathcal{C},g_i^\mathcal{N},z_i^\mathcal{C},z_i^\mathcal{N}$}
            \State {Update $\mu$ by Eqs.~(2) and (3) leveraging $z_i^\mathcal{C}$}
            \State {Get $p_i$ by Eq.~(4) leveraging $z_i^\mathcal{C}$ and $z_i^\mathcal{N}$}
            \State {Get $A_{i,j}^{(m)}$ by Eq.~(5) leveraging $p_i$
            \State {Get $A_{i,j}^{(g)}$ by Eq.~(6) leveraging $y_i^\mathcal{C}$ and $y_i^\mathcal{N}$}}
            \State { $\mathcal{L}_{\rm{gaze}}  = \frac{1}{B_{\scriptscriptstyle  C}}\sum| g_i^\mathcal{C} - y_{i}^\mathcal{C} |$}
            \State { $\mathcal{L}_{\rm{align}}^\mathcal{C}  = \frac{1}{B_{\scriptscriptstyle  C}(B_{\scriptscriptstyle  C}-1)} \sum_{x_{i}^\mathcal{C}}\sum_{x_{j}^\mathcal{C}, i\neq j} |A_{i,j}^{(g)} - A_{i,j}^{(m)}|$}
            \State {Get $A_{i,j}^{(f)}$ by Eq.~(10) leveraging $f_i^\mathcal{C}$ and $f_i^\mathcal{N}$}
            \State {$\mathcal{L}_{\rm{align}}^{\mathcal{N}}  = \frac{1}{B_{\scriptscriptstyle  N}}\sum_{x_{i}^\mathcal{N}} \frac{A_{i,:}^{(f)} \cdot A_{i,:}^{(m)}}{\| A_{i,:}^{(f)}\| \|A_{i,:}^{(m)}\|}$}
            \State {$\mathcal{L}_{\rm{all}}  = \mathcal{L}_{\rm{gaze}} + \mathcal{L}_{\rm{align}}^{\mathcal{C}}+\lambda\mathcal{L}_{\rm{align}}^{\mathcal{N}}$}
            \State {$\theta_f,\theta_g=\rm{Adam}(\mathcal{L}_{\rm{all}},\theta_{f},\theta_{g})$}
        \EndFor
        \State Calculate $\eta$ by Eq.~(7) leveraging $x_i$ and $y_i$
        \State {Repartition $\mathcal{D_S^C}$ and $\mathcal{D_S^N}$ by $\eta$}
    \EndWhile
\end{algorithmic}
\end{algorithm}

\section{Additional Quantitative Experiments}

\subsection{Experiments on the Synthetic Noisy Dataset}

First, we provide an explanation of the rationale underlying the synthetic noisy dataset design. In tasks involving noisy labels, simultaneously obtaining both noisy labels and truly accurate annotations from real-world datasets is often challenging, thereby limiting the comprehensive evaluation of methods. As a result, researchers typically conduct controlled experiments on artificially constructed noisy datasets. By examining the noisy labels in the Gaze360, as shown in Fig.~1 of the main paper, we found that their deviations from the true gaze directions are relatively large, and some of the directions are even completely opposite. Thus, we added Gaussian noise with a standard deviation of 60$\degree$ to randomly selected samples. According to the 3$\sigma$ principle of the Gaussian distribution, this setting can approximately cover noisy labels with different error levels. Moreover, based on the observations from previous work~\cite{song2022learning,xiao2015learning,li2017webvision,lee2018cleannet, song2019selfie} that real-world data typically contain about 8\%–38.5\% label noise, we introduced noise rates of 10\%, 20\%, and 30\% into the Gaze360 dataset to simulate similar conditions.

In the main paper, we have reported the results of the baseline and SeeTN on the synthetic noisy Gaze360 dataset in the form of line charts. In the supplementary material, we provide the detailed numerical results of our experiments and report the additional results of the DivideMix~\cite{li2020dividemix} and SUGE~\cite{wang2024suppressing} methods. 

As shown in Tab.~\ref{tab:synthetic_noise}, the performance of the baseline decreases as the noise ratio increases. Besides, while DivideMix underperforms on the raw Gaze360 dataset, it remains competitive against the baseline under higher noise ratios. These results suggest that although DivideMix is inherently incompatible with regression tasks, it still performs effectively under relatively high noise ratios. SUGE is specifically designed to address noisy samples in gaze estimation tasks, and it demonstrates good performance improvements across different noise ratio. However, SUGE shows a considerable performance drop when the noise ratio reaches 30\%. Our proposed SeeTN consistently achieves superior performance across all settings, and unlike SUGE, it maintains strong robustness even under a 30\% noise ratio. 

In summary, the aforementioned results emphasize the significance of noise mitigation in enhancing domain generalization, and further validate the effectiveness and robustness of SeeTN.

\subsection{Effective of Noise Handling}

To further verify that the performance improvement of our method mainly stems from its explicit modeling of noise, we train SeeTN and AGG~\cite{bao2024feature} on a clean validation set and inject 10–30\% synthetic label noise. The results are shown in Tab.~\ref{tab:synthetic_noise2}. While both SeeTN and AGG perform similarly under clean labels, SeeTN degrades much more slowly as noise increases, confirming that its gains arise from explicit noise modeling rather than domain generalization alone.

\subsection{Fine-tuned Results on the Target Domain}

\begin{table}[t]
\centering
\small
\caption{Comparison of fine-tuned results between SeeTN and state-of-the-art domain generalization methods.}
\setlength{\tabcolsep}{3pt}
\begin{tabular}{l|cc|cc}
\toprule[1pt]
Method &
$\mathcal{D}_E \!\to\! \mathcal{D}_M$ &
$\mathcal{D}_E \!\to\! \mathcal{D}_D$ &
$\mathcal{D}_G \!\to\! \mathcal{D}_M$ &
$\mathcal{D}_G \!\to\! \mathcal{D}_D$ \\
\midrule
PureGaze   & 5.30 & 6.42 & 5.20 & 7.36 \\
GazeCF  & 4.76 & 5.43 & 5.34 & 5.66 \\  
CGaG    & \textbf{4.59}  & \textbf{5.07}  & 5.13 & 5.71 \\
\midrule
SeeTN  & 5.04 & 5.23 & \textbf{4.66} & \textbf{5.52}  \\
\bottomrule[1pt]
\end{tabular}
\label{tab:compare_finetune}
\end{table}

In Tab.~\ref{tab:compare_finetune}, we compare SeeTN with several previous works~\cite{cheng2022puregaze, xu2023learning, xia2025collaborative} by fine-tuning on 100 randomly selected samples from the target domain. SeeTN achieves better performance on $\mathcal{D}_G \to \mathcal{D}_M$ and $\mathcal{D}_G \to \mathcal{D}_D$.





\subsection{More Ablation}

\begin{table}[t]
\centering
\small
\caption{The ablation of the key components in SeeTN.}
\setlength{\tabcolsep}{2pt}
\begin{tabular}{l|cc|cc}
\toprule[1pt]
Ablation &
$\mathcal{D}_E \!\to\! \mathcal{D}_M$ &
$\mathcal{D}_E \!\to\! \mathcal{D}_D$ &
$\mathcal{D}_G \!\to\! \mathcal{D}_M$ &
$\mathcal{D}_G \!\to\! \mathcal{D}_D$ \\
\midrule
SeeTN  & \textbf{6.58} & \textbf{7.18} & \textbf{6.57} & \textbf{7.57}  \\
\midrule
w/o prototypes & 7.88 & 8.14 & 7.64 & 8.37 \\
w/ L2 distance & 7.01  & 8.15  & 7.58  & 8.17 \\
\bottomrule[1pt]
\end{tabular}
\label{tab:more_ablation}
\end{table}

To validate the effectiveness of the key components in SeeTN, we construct two simple yet direct ablation studies. The first investigates the effectiveness of the prototype mechanism. Specifically, we remove the prototype module from SeeTN, and all prototype-related computations are directly replaced by the MLP-transformed features. The second examines the choice of similarity metric, where cosine similarity is replaced with the L2 distance. The results are shown in Tab.~\ref{tab:more_ablation}. We observe a substantial performance drop under all cross-domain settings, demonstrating the effectiveness of the proposed SeeTN design.

\subsection{Hyperparameter Analysis}

\begin{table}[t]
\centering
\small
\caption{The ablation of hyperparameter $\lambda$.}
\setlength{\tabcolsep}{1.5pt}
\begin{tabular}{cc|cc|cc}
\toprule[1pt]
Params. & Value & $\mathcal{D}_E \!\to\! \mathcal{D}_M$ &
$\mathcal{D}_E \!\to\! \mathcal{D}_D$ &
$\mathcal{D}_G \!\to\! \mathcal{D}_M$ &
$\mathcal{D}_G \!\to\! \mathcal{D}_D$ \\
\midrule
 & 0.01 &  6.88 & 7.41  & 6.93 & 7.74 \\
$\lambda$    & 0.1 & \textbf{6.58} & \textbf{7.18} & \textbf{6.57} & \textbf{7.57} \\
    & 1 & 7.02 & 7.66 & 6.85 & 7.70 \\
\bottomrule[1pt]
\end{tabular}
\label{tab:lambda}
\end{table}

In the Tab.~5 of the main paper, we report the ablation experiments of key parameters: $K$ and $t$. In the supplementary materials, we provide ablation experiments about the parameter $\lambda$ which controls the weight of $\mathcal{L}_{\rm{align}}^{\mathcal{N}}$, as shown in Tab.~\ref{tab:lambda}. The performance remains stable with different values of $\lambda$. Thus, we choose $\lambda=0.1$ in the SeeTN.

\section{Additional Visualization Results}

\begin{figure*}[t]
    \centering
    \includegraphics[width=\linewidth,keepaspectratio]{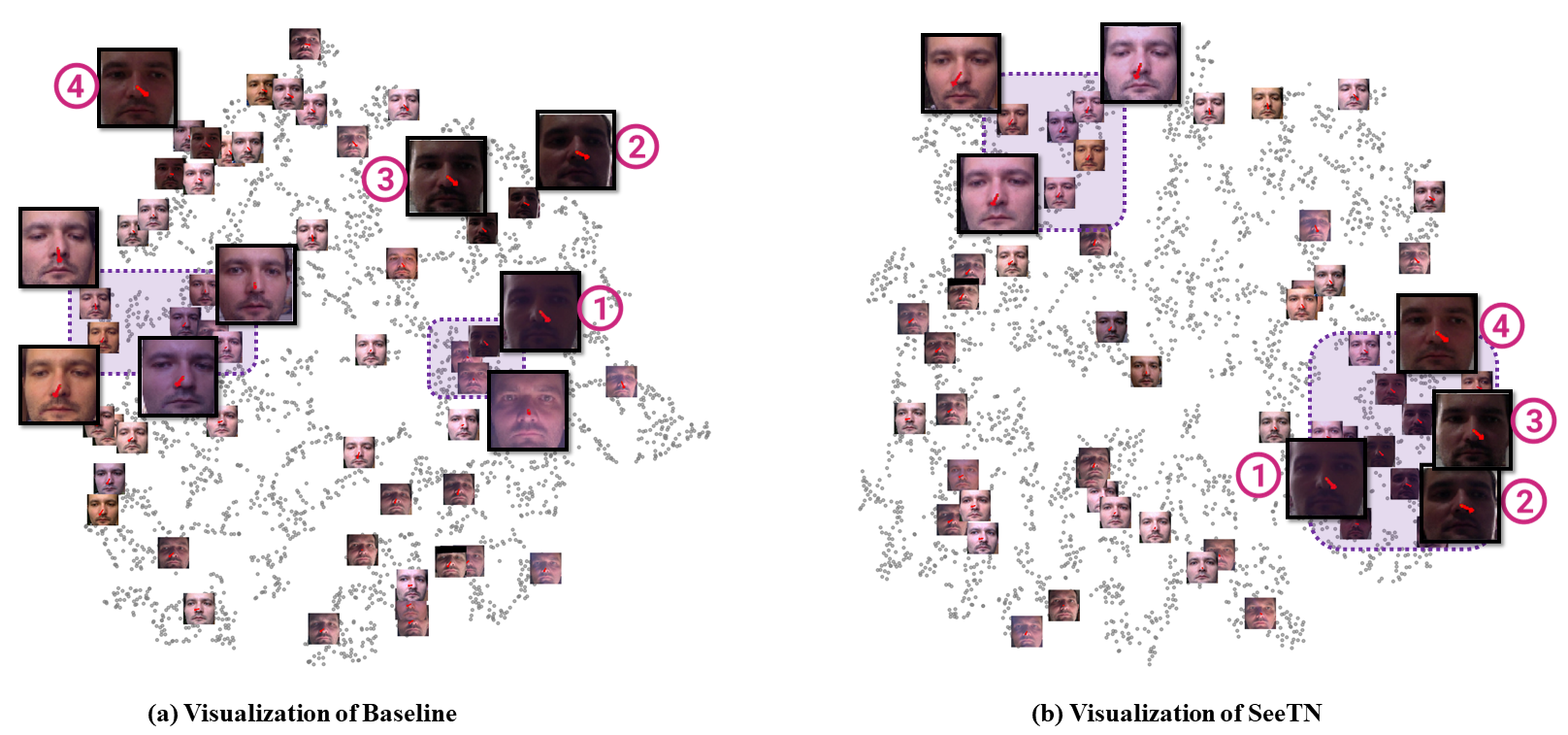}
    \caption{The qualitative results of visualizing the features learned by the backbone and SeeTN on the MPIIGaze dataset via t-SNE. In (a), samples with different gaze directions cluster together in the purple-shaded region, while in (b), samples with similar gaze directions form compact clusters. For instance, the four labeled samples, which share similar gaze directions, are dispersed in (a) but closely grouped in (b).}
    \label{fig:tsne_mpii}
\end{figure*}

\subsection{Visualization of Unseen Domain}
In tha main paper, we have shown the feature visualization of our SeeTN in source domain. To further demonstrate the generalizable abilities, we provide the t-SNE visualization results of baseline and SeeTN on the unseen domain MPIIGaze, as illustrated in Fig.~\ref{fig:tsne_mpii}. It can be observed that the baseline, shown in Fig.~\ref{fig:tsne_mpii}(a), sometimes clusters samples with different gaze directions together, whereas SeeTN, shown in Fig.~\ref{fig:tsne_mpii}(b), is more likely to group samples with similar gaze directions.  For instance, in the purple-shaded region of (a), samples with different gaze directions cluster together using baseline model, while in (b), our method can enforce the samples with similar gaze directions form compact clusters. Moreover, the four labeled samples, which share similar gaze directions, are dispersed in (a) but closely grouped in (b).

\subsection{Visualization of Noisy Sample Detection}

\begin{figure*}[!t]
    \centering
    \subfloat[Noisy Samples Selected from Gaze360]{
    \includegraphics[width=0.45\linewidth,keepaspectratio]{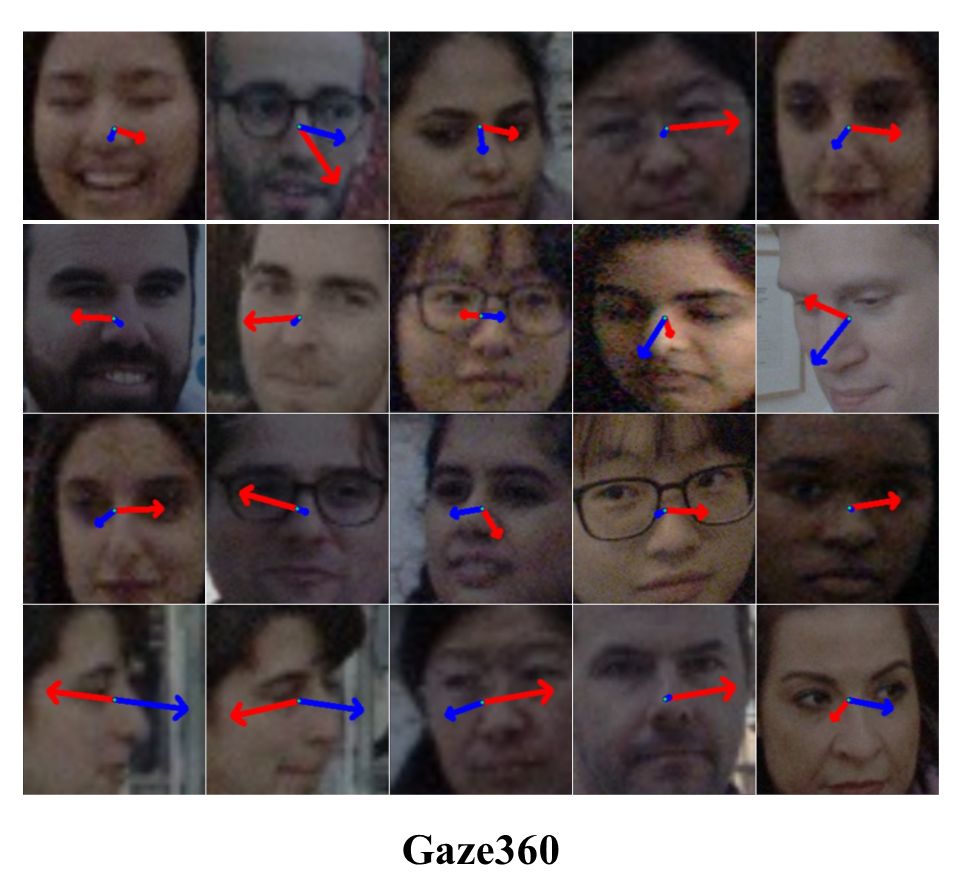}}
    \subfloat[Noisy Samples Selected from ETH-XGaze]{
    \includegraphics[width=0.45\linewidth,keepaspectratio]{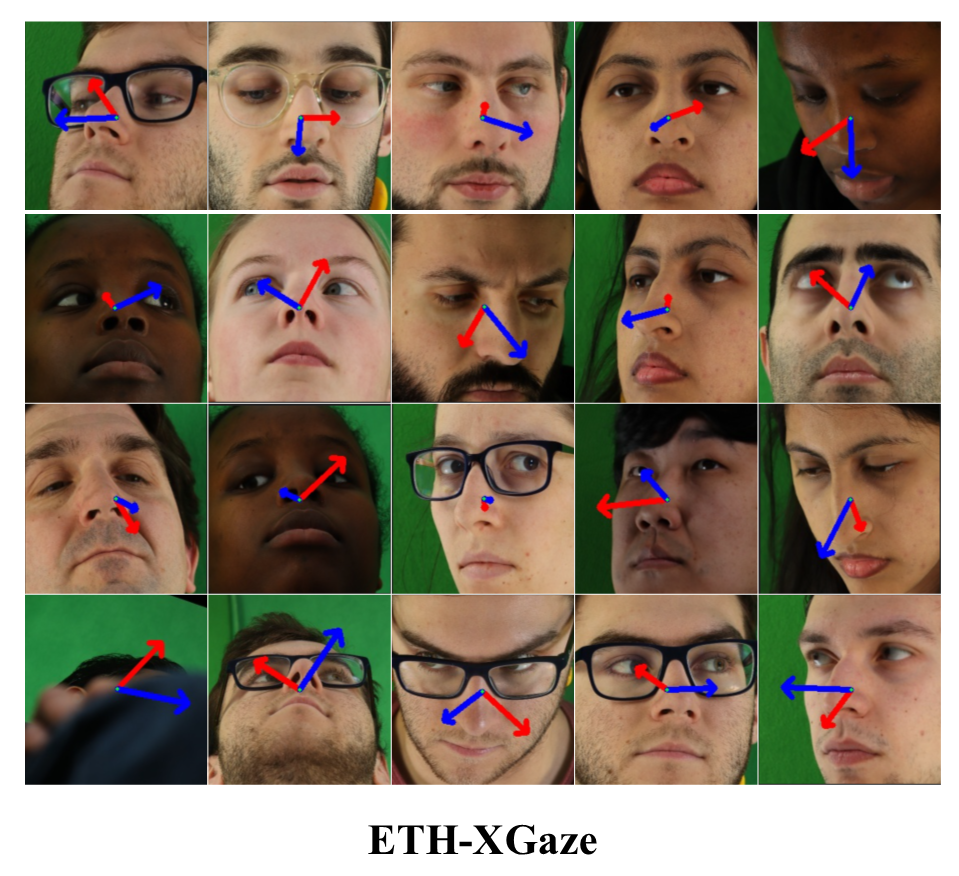}}

    \caption{Noisy Sample Visualization in Gaze360 and ETH-XGaze. The red and blue arrows indicate the ground-truth directions and the model predictions respectively. The samples selected by the our designed indicator show clear evidence of label noise.}
    \label{fig:noisy_visual}
\end{figure*}

We present additional visualizations of noisy samples selected by our proposed indicator $\eta$, as shown in Fig.~\ref{fig:noisy_visual}(a) and Fig.~\ref{fig:noisy_visual}(b). We can see that the ground-truth labels of our selected samples are largely incorrect.

\begin{figure*}[!t]
    \centering
    \subfloat[Noisy Samples from MPIIGaze]{
    \includegraphics[width=0.45\linewidth,keepaspectratio]{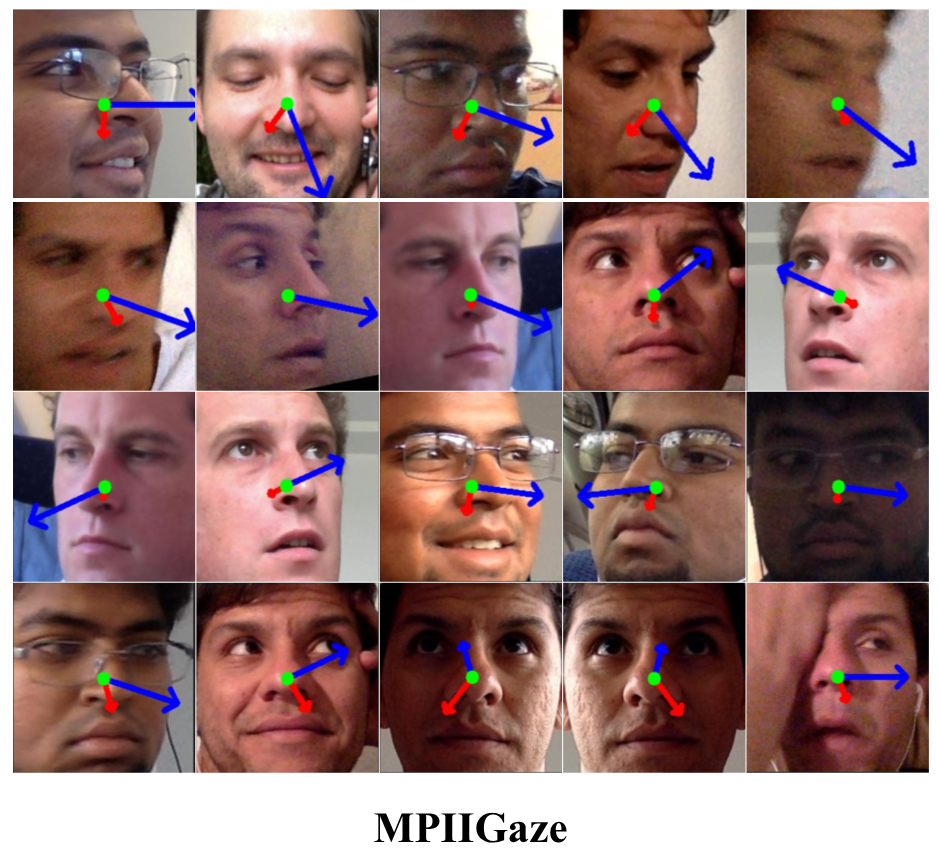}}
    \subfloat[Noisy Samples from EyeDiap]{
    \includegraphics[width=0.45\linewidth,keepaspectratio]{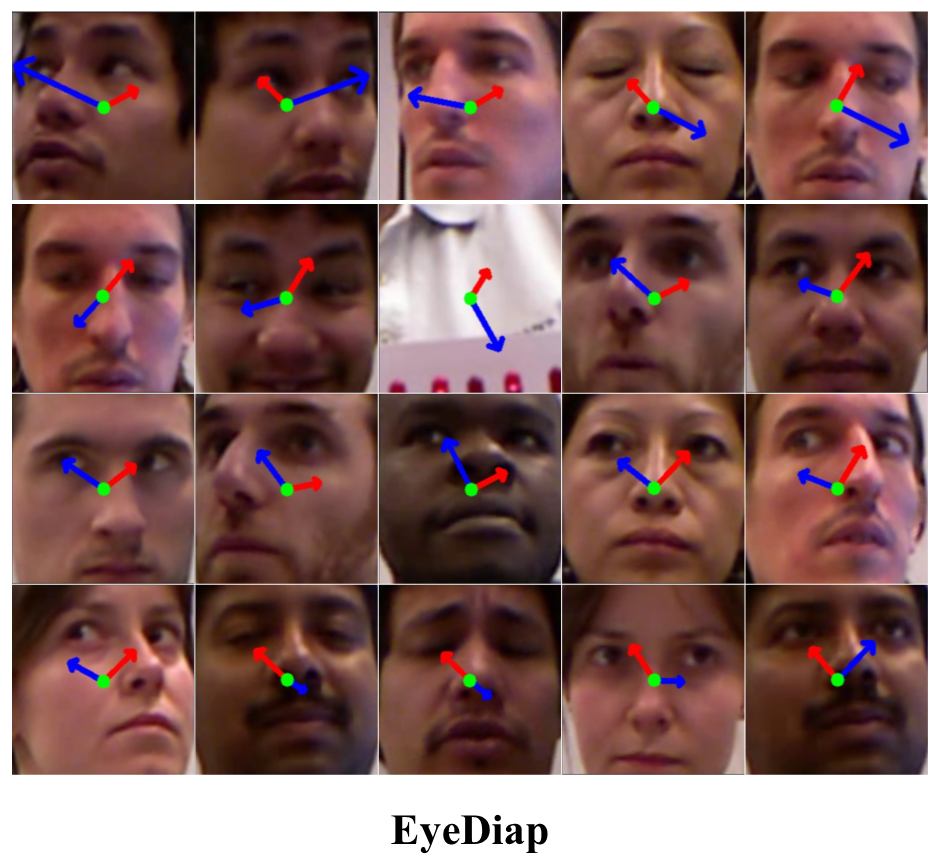}}

    \caption{Noisy Sample Visualization in MPIIGaze and EyeDiap. The red and blue arrows indicate the ground-truth directions and the model predictions respectively. It can be observed that some of samples with incorrect labels are correctly predicted by our SeeTN model in the unseen domain.}
    \label{fig:noisy_target}
\end{figure*}

Besides, we also visualize the unseen-domain samples with large test errors for SeeTN in Fig.~\ref{fig:noisy_target}. In fact, a considerable number of noisy labels are also present in the unseen domain, while SeeTN can correctly predict their truly gaze directions, which demonstrates the widespread presence of noise in gaze estimation tasks. Meanwhile, the generalization ability of gaze estimation models across domains also requires further investigation.

\end{document}